\documentclass[sigconf,screen]{acmart}
\usepackage{color}
\usepackage[ruled]{algorithm2e}
\usepackage{bm}
\usepackage{multirow}
\usepackage{booktabs}
\usepackage{balance}
\renewcommand\footnotetextcopyrightpermission[1]{} 
\def\ie{{\em i.e.}}
\def\eg{{\em e.g.}}

\newcommand{\figref}[1]{Fig. \ref{#1}}

\newcommand{\secref}[1]{Sect. \ref{#1}}

\newcommand{\mc}[1]{\mathcal{#1}}

\newsavebox\CBox

\AtBeginDocument{%
  \providecommand\BibTeX{{%
    \normalfont B\kern-0.5em{\scshape i\kern-0.25em b}\kern-0.8em\TeX}}}
\copyrightyear{2020} 
\acmYear{2020} 
\setcopyright{acmcopyright}
\acmConference[MM '20]{Proceedings of the 28th ACM International Conference on Multimedia}{October 12--16, 2020}{Seattle, WA, USA}
\acmBooktitle{Proceedings of the 28th ACM International Conference on Multimedia (MM '20), October 12--16, 2020, Seattle, WA, USA}
\acmPrice{15.00}
\acmDOI{10.1145/3394171.3413946}
\acmISBN{978-1-4503-7988-5/20/10}
\begin{document}
\fancyhead{}
\title{Cooperative Bi-path Metric for Few-shot Learning}

\author{Zeyuan Wang$^{1}$, Yifan Zhao$^{1}$, Jia Li$^{1,3,4*}$, Yonghong Tian$^{2,4}$}
\thanks{$^{*}$Jia Li is the corresponding author (E-mail: \textsuperscript{}jiali@buaa.edu.cn).\\ Website: \url{http://cvteam.net}}
\affiliation{%
    \institution{$^1$State Key Laboratory of Virtual Reality Technology and Systems, SCSE, Beihang University, Beijing, 100191, China}
    \institution{$^2$School of Electronics Engineering and Computer Science, Peking University, Beijing, 100871, China}
    \institution{$^3$Beijing Advanced Innovation Center for Big Data and Brain Computing, Beihang University, Beijing, 100191, China}
    \institution{$^4$Peng Cheng Laboratory, Shenzhen, 518066, China}
}

\renewcommand{\shortauthors}{Wang and Zhao, et al.}
\begin{abstract}
    Given base classes with sufficient labeled samples, the target of few-shot classification is to recognize unlabeled samples of novel classes with only a few labeled samples. 
    Most existing methods only pay attention to the relationship between labeled and unlabeled samples of novel classes, which do not make full use of information within base classes. In this paper, we make two contributions to investigate the few-shot classification problem. First, we report a simple and effective baseline trained on base classes in the way of traditional supervised learning, which can achieve comparable results to the state of the art. Second, based on the baseline, we propose a cooperative bi-path metric for classification, which leverages the correlations between base classes and novel classes to further improve the accuracy. Experiments on two widely used benchmarks show that our method is a simple and effective framework, and a new state of the art is established in the few-shot classification field.
\end{abstract}

\maketitle
\section{Introduction}
\begin{figure}[]
    \includegraphics[width=\linewidth]{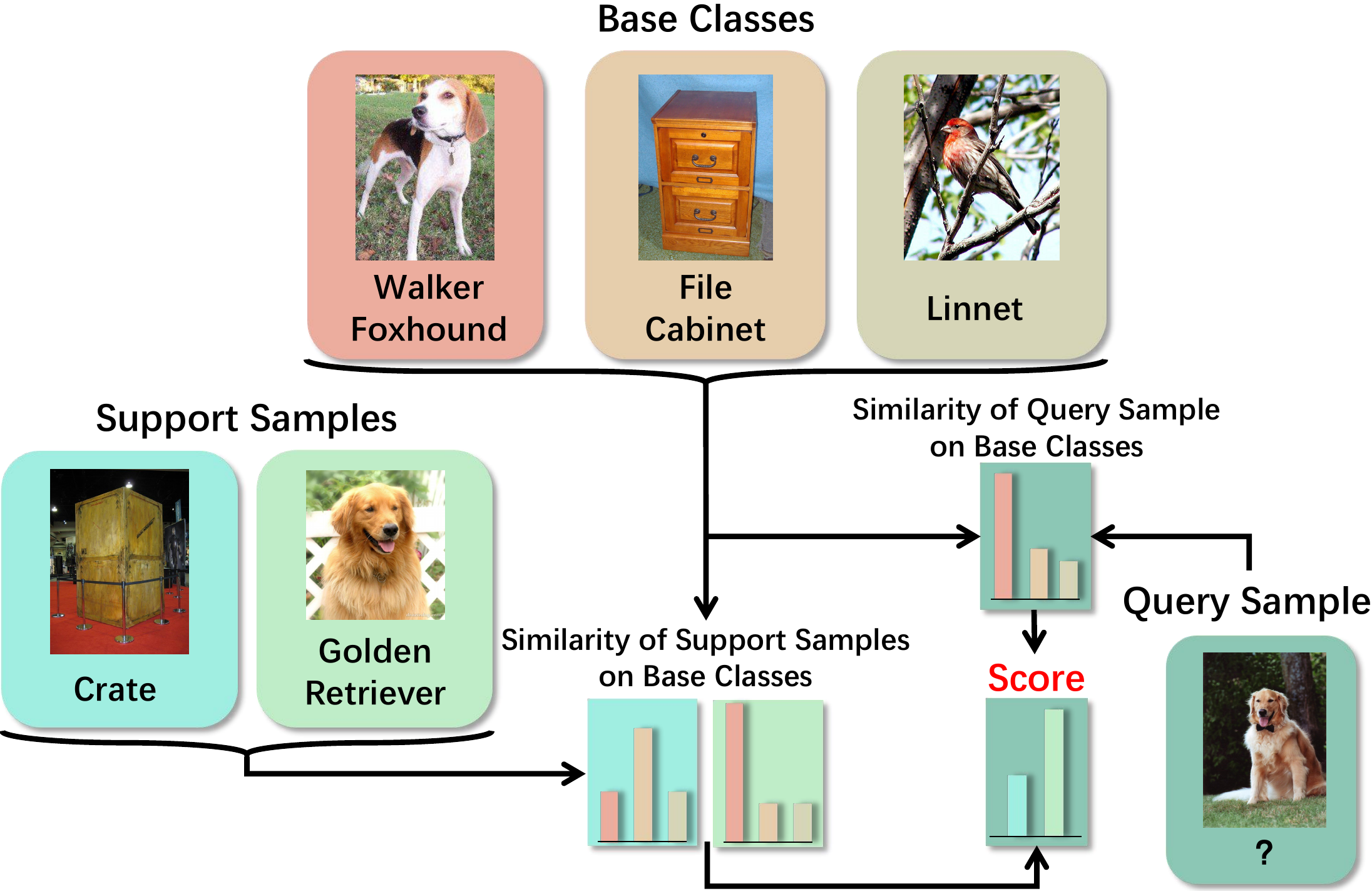}
    \caption{Motivation of the proposed approach.}
    \label{fig:1}
\end{figure}

Image recognition provides an intuitive and fundamental way to understand the visual world, which is also deeply rooted in many advanced kinds of research, including autonomic production and artificial intelligence. As the crucial part of visual content, image recognition has witnessed a great progress with the proposals of large dataset~\cite{DBLP:journals/ijcv/RussakovskyDSKS15} and deep learning techniques~\cite{DBLP:conf/nips/KrizhevskySH12,DBLP:conf/cvpr/HeZRS16}. However, the notable performance of these methods heavily relies on the large amount of manually annotated datasets, which are labour-intensive and sometimes inaccessible,~\eg, ImageNet~\cite{DBLP:journals/ijcv/RussakovskyDSKS15} with over 15 million annotations. Remarkably, humans and other animals seem to have the potential to recognize one identity with very less related knowledge. Therefore, the few-shot learning (FSL) problem with limited seen knowledge forces the model to make a typical generalization of each class, which is a more realistic setting in some extreme industrial applications.

By training a model on base classes that contain sufficient labeled samples, the goal of few-shot learning is to make the model generalize well on the novel classes which do not intersect with the base classes, \emph{i.e.}, correctly classifying unlabeled samples (\emph{query samples}) according to a small number of labeled samples (\emph{support samples}). To make the conditions of the training phase match those of the testing phase, Matching Networks \cite{DBLP:conf/nips/VinyalsBLKW16} firstly suggested that both training and testing should adopt \emph{episodic} procedure, which came from meta-learning. Models will meet many few-shot learning tasks in both the training and testing phases. Each task consists of several classes, and each class contains a few support samples and several query samples. Leading by the pioneer work \cite{DBLP:conf/nips/VinyalsBLKW16}, many subsequent researches \cite{DBLP:journals/corr/LiZCL17,DBLP:conf/iclr/RaviL17,DBLP:conf/icml/FinnAL17,DBLP:conf/nips/TriantafillouZU17,DBLP:conf/nips/SnellSZ17,DBLP:journals/corr/abs-1803-02999,DBLP:conf/cvpr/SungYZXTH18,DBLP:conf/iclr/RenTRSSTLZ18,DBLP:conf/iclr/MishraR0A18,DBLP:conf/nips/OreshkinLL18,DBLP:conf/cvpr/LiWXHGL19,DBLP:conf/iccv/Qiao000HW19,DBLP:conf/nips/HouCMSC19} followed this episodic learning and achieved notable improvements. However, some recent works \cite{DBLP:conf/cvpr/GidarisK18,DBLP:conf/cvpr/QiaoLSY18,DBLP:conf/cvpr/LifchitzAPB19,DBLP:journals/corr/abs-1909-02729} did not follow this factitious sampling setting, but directly trained model in the way of traditional supervised learning. Thus a natural concern arises, \textbf{is episodic training necessary for few-shot learning?} Keeping this in our mind, we first make extensive experimental analyses on two commonly used benchmarks \cite{DBLP:conf/nips/VinyalsBLKW16,DBLP:conf/iclr/RenTRSSTLZ18}. Counter-intuitively, without the always-used episodic training process, state-of-the-art performance can also be achieved through adequate training strategy with all the samples in base classes. \textbf{This finding not only brings us a rethinking of this conventional setting but also can be considered as a high-performance baseline for few-shot learning.}
\begin{figure}[]
    \includegraphics[width=\linewidth]{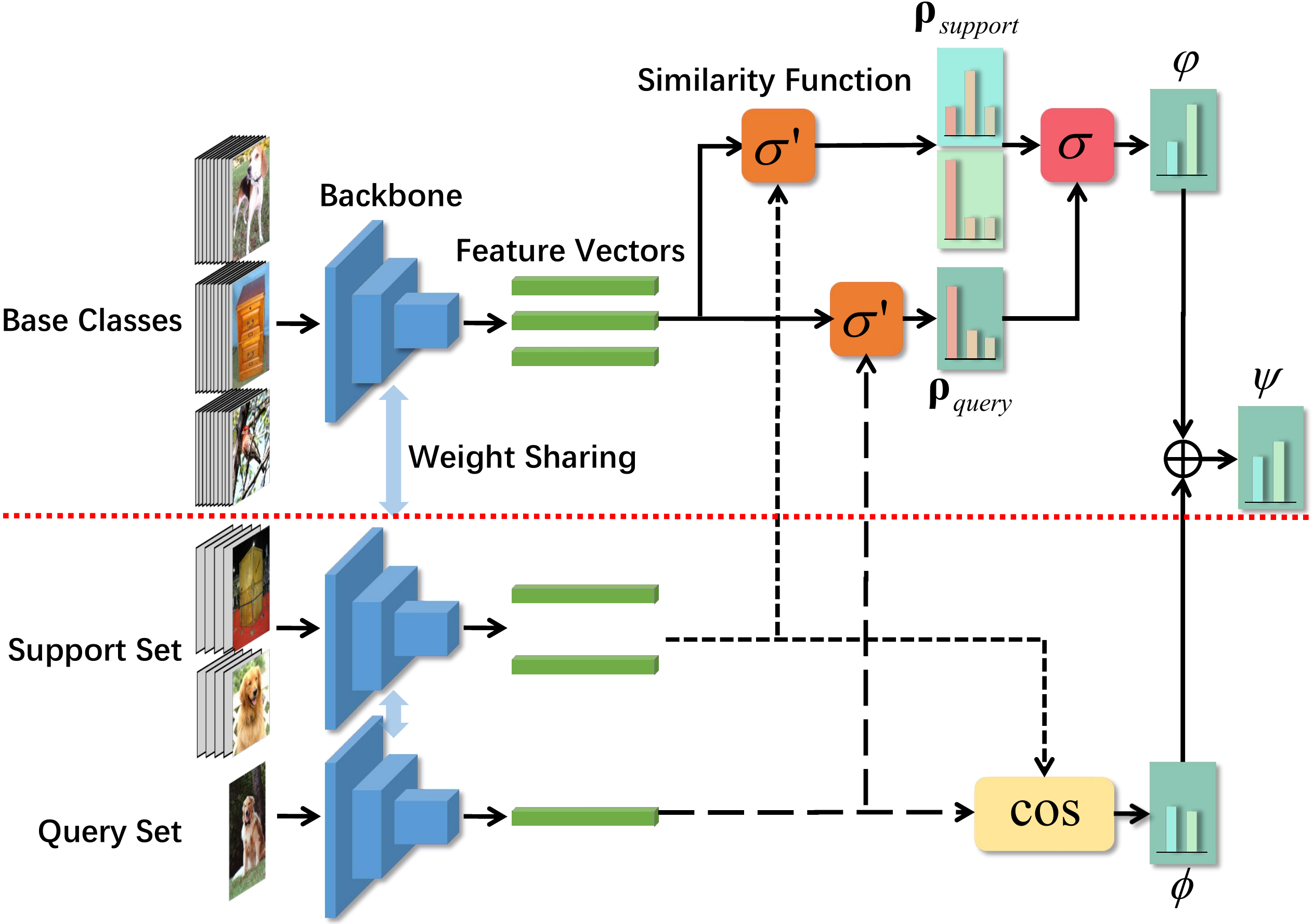}
    \caption{Illustration of the pipeline of Cooperative Bi-path Metric. The top half part upon the \textcolor[RGB]{255,0,0}{red} dotted line represents our proposed transductive similarity, and the bottom half part is the classic inductive similarity. ${\bm{\uprho }}_\textit{support}$ refers to the similarity distributions of support set on base classes, and ${\bm{\uprho }}_\textit{query}$ refers to those of query set. The final classification score $\psi$ is the weighted sum of ${\phi}$ and ${\varphi}$, where ${\phi}$ is the conventional cosine similarity between support set and query set, and ${\varphi}$ is the similarity between ${\bm{\uprho }}_\textit{support}$ and ${\bm{\uprho }}_\textit{query}$.}
    \label{fig:2}
\end{figure}

To start from another perspective, metric learning \cite{DBLP:journals/air/AtkesonMS97,DBLP:conf/nips/GoldbergerRHS04,DBLP:conf/nips/Fink04,DBLP:conf/cvpr/ChopraHL05,DBLP:conf/nips/WeinbergerBS05,DBLP:journals/jmlr/SalakhutdinovH07,DBLP:journals/jmlr/WeinbergerS09,DBLP:conf/cvpr/KostingerHWRB12,DBLP:journals/corr/BelletHS13,DBLP:journals/ftml/Kulis13,DBLP:journals/corr/HofferA14,DBLP:conf/icml/HarandiSH17} is a major genre in the field of few-shot learning. This kind of method classifies query samples by learning a feature extractor on the base classes, extracting features of samples of the novel classes during testing, and measuring the distance or similarity between labeled support samples and unlabeled query samples. However, most of the existing metric learning methods \cite{DBLP:conf/nips/VinyalsBLKW16,DBLP:conf/nips/SnellSZ17,DBLP:conf/cvpr/SungYZXTH18,DBLP:conf/nips/OreshkinLL18,DBLP:conf/cvpr/LiWXHGL19,DBLP:conf/cvpr/LifchitzAPB19,DBLP:conf/iccv/Qiao000HW19,DBLP:conf/nips/HouCMSC19} for few-shot learning focused on the correlation between support samples and query samples within novel classes, and did not take full use of the information of base classes. Unlike the above-mentioned approaches, we make use of the base classes to assist in classifying query samples of novel classes during the testing phase. \textbf{Our motivation is that for samples with identical labels of novel classes, their similarity distributions on base classes are more likely to be similar, while different classes' samples should have different distributions from each other.} For example, as shown in the Fig.~\ref{fig:1}, the \textit{golden retrievers} are more similar to \textit{walker foxhounds} than \textit{file cabinets}, meanwhile the \textit{crates} are more similar to \textit{file cabinets} than \textit{walker foxhounds}. We emphasize this kind of relationship information between the base classes and the novel classes is valuable and helpful for classifying unlabeled samples. Inspired by this finding, we propose a novel metric named \emph{Cooperative Bi-path Metric}, as shown in Fig.~\ref{fig:2}, which not only considers the inductive similarity between support set and query set in the novel classes but also measures their transductive similarity distributions through base classes and further takes \textbf{the similarity of their similarity distributions on base classes} into consideration.

Our main contribution is three-fold:
1) We make extensive experimental analyses of the conventional episodic training and full-supervision training for the few-shot classification problem, and propose a new high-performance baseline based on this experimental finding.
2) We propose a novel Cooperative Bi-path Metric learning approach, which first exploits the base classes as an intermediary for facilitating the classification process.
3) We make extensive experimental analyses to demonstrate our findings. Compared with the existing methods, our proposed approach achieves new state-of-the-art results on two widely used benchmarks,~\ie, miniImageNet \cite{DBLP:conf/nips/VinyalsBLKW16} and tieredImageNet \cite{DBLP:conf/iclr/RenTRSSTLZ18}.

\section{Related work}
There are two main branches of methods in the few-shot learning field, one is meta-learning based methods and the other is metric-learning based methods. The former adopts the episodic training procedure in the training stage and expects to learn the common attributes between different tasks through this procedure, namely the appropriate hyperparameters for models. The latter focuses on how to better extract features from samples and classify samples according to the extracted features, that is, the network is expected to learn a function that can properly measure the similarity between samples.

\noindent \textbf{Meta-learning based methods.} The purpose of meta-learning, or learning to learn \cite{DBLP:books/sp/98/Thrun98,DBLP:books/sp/98/ThrunP98,DBLP:journals/air/VilaltaD02}, is to train a \emph{meta-learner} to learn task-agnostic knowledge (or hyper-parameters), which can assist the training of learners on different tasks. Meta-learning based methods \cite{DBLP:conf/iclr/RaviL17,DBLP:journals/corr/LiZCL17,DBLP:conf/icml/FinnAL17,DBLP:journals/corr/abs-1803-02999,DBLP:conf/cvpr/GidarisK18,DBLP:conf/cvpr/QiaoLSY18} are a major branch in the field of few-shot learning. For example, Meta-LSTM \cite{DBLP:conf/iclr/RaviL17} trained an LSTM-based \cite{DBLP:journals/neco/HochreiterS97} meta-learner to discover good initialization for learner's parameters, as well as a mechanism for updating the learner's parameters by a small sample set. Similarly, Meta-SGD \cite{DBLP:journals/corr/LiZCL17} trained a meta-learner to produce learner's initialization, update direction, and learning rate, but in a single meta-learning process. Also, MAML \cite{DBLP:conf/icml/FinnAL17} aimed to find the appropriate initial parameters of the learner, so that the learner could converge rapidly with a few samples. Similar to MAML \cite{DBLP:conf/icml/FinnAL17}, Reptile \cite{DBLP:journals/corr/abs-1803-02999} was a meta-learning method for finding neural network initialization parameters but simply performed SGD on each task without computing gradient twice as MAML \cite{DBLP:conf/icml/FinnAL17} did. Gidaris \emph{et al.} \cite{DBLP:conf/cvpr/GidarisK18} and Qiao \emph{et al.} \cite{DBLP:conf/cvpr/QiaoLSY18} both used base classes to train a feature extractor in the first stage, then used the base classes training a classification weight generator (meta-learner) in the second stage, and used the classification weight generator and support samples to generate classification weight vectors (parameters of the learner) for novel classes during the testing phase. As we can see, all the above meta-learning methods require training a meta-learner and more or less fine-tuning on novel classes, which undoubtedly increases the complexity of the methods. However, our baseline is trained in the way of supervised learning, which can achieve comparable performance to the state of the art, meanwhile greatly simplifies the training process.

\noindent \textbf{Metric-learning based methods.} Another branch in the few-shot learning field  is metric-learning based methods \cite{DBLP:conf/nips/VinyalsBLKW16,DBLP:conf/nips/SnellSZ17,DBLP:conf/nips/TriantafillouZU17,DBLP:conf/cvpr/SungYZXTH18,DBLP:conf/iclr/RenTRSSTLZ18,DBLP:conf/nips/OreshkinLL18}, which focus on embedding samples into a metric space so that the samples can be classified according to similarity or distance between each other. Matching Networks \cite{DBLP:conf/nips/VinyalsBLKW16} used LSTM \cite{DBLP:journals/neco/HochreiterS97} and variants to extract feature vectors from support samples and query samples, and then classified query samples by calculating cosine similarities between support and query samples. Prototypical Networks \cite{DBLP:conf/nips/SnellSZ17} took the average of the feature vectors of support samples within a class as the prototype of the class, and assigned the query samples to the nearest prototype in Euclidean distance,  meanwhile its convolutional neural network was trained end-to-end. Triantafillou \emph{et al.} \cite{DBLP:conf/nips/TriantafillouZU17} proposed an information retrieval-inspired approach viewing each batch point as a query that ranked the remaining ones and defined a model to optimize mean Average Precision over these rankings. Relation Network \cite{DBLP:conf/cvpr/SungYZXTH18} adapted the same convolutional neural network as Prototypical Networks \cite{DBLP:conf/nips/SnellSZ17} to extract the features of support samples and query samples, but features of support samples and query samples were concatenated and input into nonlinear relation module to obtain classification scores. Ren \emph{et al.} \cite{DBLP:conf/iclr/RenTRSSTLZ18} presented a new problem: semi-supervised few-shot classification, in which support samples consisted of labeled and unlabeled samples, and proposed several novel extensions on Prototypical Networks \cite{DBLP:conf/nips/SnellSZ17}. Also, TADAM \cite{DBLP:conf/nips/OreshkinLL18} learned a task-dependent metric with Metric Scaling, Task Conditioning, and Auxiliary Task Co-training for few-shot classification. Although the above metric-learning based methods could achieve admirable results on few-shot classification problem, they only considered the direct relationship between query samples and support samples when predicting the labels of query samples of novel classes, and ignored the relationship between novel classes and base classes. After observing this, we propose a novel Cooperative Bi-path Metric to utilize the relationship information between the support samples, query samples, and base classes to assist in the classification of queries, which can further improve the accuracy of classification.
\section{method}
\subsection{Problem Definition}
Given a training set $\mathbb{D}{_\textit{base}}$ containing samples of base classes $\mathbb{C}{_\textit{base}}$, the goal of few-shot learning is to train a model with $\mathbb{D}{_\textit{base}}$ to achieve high accuracy on the classification tasks obtained by sampling on test set $\mathbb{D}{_\textit{novel}}$, which contains samples of novel classes $\mathbb{C}{_\textit{novel}}$. The base classes are totally different from novel classes, that is, $\mathbb{C}{_\textit{base}}\cap\mathbb{C}{_\textit{novel}}=\varnothing$. And each task is composed of a support set $\mathbb{D}{_\textit{support}}$ with labeled samples and a query set $\mathbb{D}{_\textit{query}}$ with unlabeled samples. For a \emph{N}-way \emph{K}-shot task, ${\mathbb{D}_\textit{support}} = \left\{ {\left( {{{\mathbf{x}}^{\left( i \right)}},{y^{\left( i \right)}}} \right)} \right\}_{i = 1}^{N \times K}$ contains \emph{N} classes and \emph{K} support samples for each class. A model is trained to predict the labels of the query samples in $\mathbb{D}{_\textit{query}}$ as accurately as possible according to $\mathbb{D}{_\textit{support}}$.


\subsection{A Strong Baseline for Few-shot Learning}
\label{sec:3.2}

\noindent \textbf{Is episodic training necessary?}
In the view of episodic training, a lot of works \cite{DBLP:conf/nips/VinyalsBLKW16,DBLP:journals/corr/LiZCL17,DBLP:conf/iclr/RaviL17,DBLP:conf/icml/FinnAL17,DBLP:conf/nips/TriantafillouZU17,DBLP:conf/nips/SnellSZ17,DBLP:journals/corr/abs-1803-02999,DBLP:conf/cvpr/SungYZXTH18,DBLP:conf/iclr/RenTRSSTLZ18,DBLP:conf/iclr/MishraR0A18,DBLP:conf/nips/OreshkinLL18,DBLP:conf/cvpr/LiWXHGL19,DBLP:conf/iccv/Qiao000HW19,DBLP:conf/nips/HouCMSC19} made the procedure of training and testing consistent to achieve higher performance. During training phase, tasks were sampled, then loss function of the model was calculated based on the tasks, and the network parameters were updated through the back propagation. However, some works \cite{DBLP:conf/cvpr/GidarisK18,DBLP:conf/cvpr/QiaoLSY18,DBLP:conf/cvpr/LifchitzAPB19,DBLP:journals/corr/abs-1909-02729} did not follow this setting, but trained classification networks in the way of traditional supervised learning.

To make fair comparisons with the previous works \cite{DBLP:conf/iclr/MishraR0A18,DBLP:conf/iccv/Qiao000HW19,DBLP:conf/nips/OreshkinLL18,DBLP:conf/cvpr/LifchitzAPB19,DBLP:conf/nips/HouCMSC19}, we adopt ResNet-12 \cite{DBLP:conf/cvpr/HeZRS16} as the baseline's backbone and use the training set $\mathbb{D}{_\textit{base}}$ to optimize it in a fully-supervised manner. However, it is interesting to find that a network with only a supervised training strategy can also get superior classification results on novel classes (elaborated in~\secref{sec:4}). This indicates the basic knowledge learnt from base classes can not be further improved by the episodic learning and motivates the proposal of our strong baseline.

\noindent \textbf{Baseline for few-shot learning.}
For the problem of few-shot learning, we advocate three meaningful cues in constructing a strong baseline:
1) Data augmentation: following prior work\cite{DBLP:conf/nips/HouCMSC19}, we use horizontal flip, random crop and random erasing \cite{DBLP:journals/corr/abs-1708-04896} as data augmentation.
2) Temperature in learning: inspired by \cite{DBLP:journals/corr/HintonVD15}, we also introduce a hyper-parameter called temperature, which was first applied in model distillation to change the smoothness of distribution after softmax normalization and the value of cross-entropy.
3) Dense classification: instead of embedding the image features as a vector, we apply dense classification loss \cite{DBLP:conf/cvpr/LifchitzAPB19} to regularize our model, \emph{i.e.}, all the local feature vectors of the feature map before the last fully-connected layer are classified through the fully-connected layer without average pooling. For each training sample $\left( {{\mathbf{x}},y} \right) \in {\mathbb{D}_\textit{base}}$, the proposed baseline with loss $\mc{L}$ has the following form:
\begin{gather}
    \mc{L} =  - \frac{1}{r}\sum\limits_{i = 1}^r {\log \frac{{\exp \left( {t\left( {{{\mathbf{f}}^{\left( i \right) \top }}{{\mathbf{p}}^{\left( y \right)}} + {{\mathbf{b}}_y}} \right)} \right)}}{{\sum\limits_{j = 1}^{\left| {{\mathbb{C}_\textit{base}}} \right|} {\exp \left( {t\left( {{{\mathbf{f}}^{\left( i \right) \top }}{{\mathbf{p}}^{\left( j \right)}} + {{\mathbf{b}}_j}} \right)} \right)} }}},
\end{gather}
where $t$ is the temperature hyperparameter. $\left|  \cdot  \right|$ is the cardinality of a set. ${{{\mathbf{f}}^{\left( i \right)}}} \in {\mathbb{R}^c}$ is the local vector at position $i$ of the training sample's feature map ${\mathbf{F}} \in {\mathbb{R}^{c \times r}}$ with channel dimension $c$ and spatial resolution $r$. ${{{\mathbf{p}}^{\left( j \right)}}} \in {\mathbb{R}^c}$ is the parameter vector for class $j$ in the fully-connected layer's parameter matrix $\mathbf{P} \in {\mathbb{R}^{c \times {\left| {{\mathbb{C}_\textit{base}}} \right|}}}$. ${{{\mathbf{b}}_j}}$ is the bias for class $j$ in the fully-connected layer's bias vector ${\mathbf{b}} \in {\mathbb{R}^{{\left| {{\mathbb{C}_\textit{base}}} \right|}}}$.

During testing phase, for a \emph{N}-way \emph{K}-shot task, a query sample in $\mathbb{D}{_\textit{query}}$ is assigned to the class $\hat y$ with maximum classification score ${{\phi ^{\left( n \right)}}}$:
\begin{gather}
    \label{eq:2}
    \hat y = \mathop {\arg \max }\limits_n \left( {{\phi ^{\left( n \right)}}} \right).
\end{gather}
Classification score ${{\phi ^{\left( n \right)}}}$ for novel class $n$ is defined as:
\begin{gather}
    \label{eq:3}
    {\phi ^{\left( n \right)}} = \cos \left( {{\mathbf{q}},{{\mathbf{s}}^{\left( n \right)}}} \right).
\end{gather}
And $\cos \left( , \right)$ is cosine similarity between two vectors. ${\mathbf{q}}$ and ${{\mathbf{s}}^{\left( n \right)}}$ are feature vectors of query sample and class $n$ respectively:
\begin{equation}
    \begin{gathered}
        \label{eq:4}
        \cos \left( {{\mathbf{a}},{\mathbf{b}}} \right) = \frac{{{{\mathbf{a}}^ \top }{\mathbf{b}}}}{{\left\| {\mathbf{a}} \right\| \left\| {\mathbf{b}} \right\|}},\\
        {\mathbf{q}} = GAP\left( {\mathbf{Q}} \right),\\
        {{\mathbf{s}}^{\left( n \right)}} = \frac{1}{K}\sum\limits_{k = 1}^K {GAP\left( {{{\mathbf{S}}^{\left( {n,k} \right)}}} \right)}.
    \end{gathered}
\end{equation}
And $\left\| \cdot \right\|$ is the $L_\textit{2}$ norm of a vector. ${\mathbf{Q}}$ and ${{{\mathbf{S}}^{\left( {n,k} \right)}}}$ are feature maps of the query sample and the $k$-th support sample of class $n$ in $\mathbb{D}{_\textit{support}}$. $GAP\left(  \cdot  \right)$ is the global average pooling on a feature map $\mathbf{F}$ defined as:
\begin{gather}
    \label{eq:5}
    GAP\left( {\mathbf{F}} \right) = \frac{1}{r}\sum\limits_i^r {{{\mathbf{f}}^{\left( i \right)}}},
\end{gather}
where ${{{\mathbf{f}}^{\left( i \right)}}} \in {\mathbb{R}^c}$ is the local vector at position $i$ of the feature map ${\mathbf{F}} \in {\mathbb{R}^{c \times r}}$.

\subsection{Cooperative Bi-path Metric}
\label{sec:Cooperative Bi-path Metric}
As shown in Eq.~(\ref{eq:2}) and (\ref{eq:3}), the previously proposed baseline in this paper as well as previous methods \cite{DBLP:conf/nips/VinyalsBLKW16,DBLP:conf/nips/SnellSZ17,DBLP:conf/cvpr/SungYZXTH18,DBLP:conf/nips/OreshkinLL18,DBLP:conf/cvpr/LiWXHGL19,DBLP:conf/cvpr/LifchitzAPB19,DBLP:conf/iccv/Qiao000HW19,DBLP:conf/nips/HouCMSC19} simply classify query samples only according to support samples. The main drawback is that the prior knowledge on base classes is not fully exploited, which can also be complementary to classification decision.
Thus a natural thought arises: the similarity distributions on base classes of support samples and query samples within the same class should also be similar, as illustrated in Fig.~\ref{fig:1}, and this information is useful for classifying query samples.

Starting from this point, we propose a novel method namely Cooperative Bi-path Metric as the classification criterion during the testing phase, which is shown in ~\figref{fig:2}. Cooperative Bi-path Metric utilizes base classes as an intermediate way to assist with the classification of query samples.
Our proposed metric measures the similarity by two individual paths: inductive similarity ${\phi}$ and transductive similarity ${\varphi}$. Most existing methods regard the former one as the only classification criterion, as shown in the lower half of Fig.~\ref{fig:2}, which calculates the inductive similarity ${\phi}$ (\emph{e.g.} cosine similarity) between the support set and the query set. While Cooperative Bi-path Metric not only measures the inductive similarity ${\phi}$ but also uses base classes as an agent to calculate the transductive similarity ${\varphi}$ between the support set and the query set, as shown in the upper part of Fig.~\ref{fig:2}. Firstly, it calculates the similarity distribution ${\bm{\uprho }}_\textit{support}$ and ${\bm{\uprho }}_\textit{query}$ of support set and query set on base classes, and then calculates the similarity between ${\bm{\uprho }}_\textit{support}$ and ${\bm{\uprho }}_\textit{query}$, \emph{i.e.}, the transductive similarity ${\varphi}$ between support set and query set. The final classification score ${\psi}$ during the test phase is a weighted sum of ${\phi}$ and ${\varphi}$:
\begin{equation}
    \begin{gathered}
        \label{eq:6}
        \hat y = \mathop {\arg \max }\limits_n \left( {\psi ^{\left( n \right)}} \right),\\
        {\psi ^{\left( n \right)}} = \alpha {\phi ^{\left( n \right)}} + \left( {1 - \alpha} \right){\varphi ^{\left( n \right)}},\\
        {\varphi ^{\left( n \right)}} = \sigma \left( {{{\bm{\uprho }}_{\textit{query}}},{\bm{\uprho }}_\textit{support}^{\left( n \right)}} \right).\\
    \end{gathered}
\end{equation}
And ${\psi ^{\left( n \right)}}$ is Cooperative Bi-path Metric's final classification score for novel class $n$. ${\phi ^{\left( n \right)}}$ is defined in Eq.~(\ref{eq:3}). $\alpha$ is a hyperparameter to adjust the weight between ${\phi ^{\left( n \right)}}$ and ${\varphi ^{\left( n \right)}}$. $\sigma \left( , \right)$ is a similarity function that measures the similarity between two distributions, and it can be cosine similarity or negative Euclidean distance and so on. ${{\bm{\uprho }}_{\textit{query}}}$ and ${\bm{\uprho }}_\textit{support}^{\left( n \right)}$ can be formally represented as:
\begin{equation}
    \begin{gathered}
        \label{eq:7}
        {{\bm{\uprho }}_{\textit{query}}} = \sigma '\left( {{\mathbf{q}},{\mathbf{B}}} \right),\\
        {\bm{\uprho }}_\textit{support}^{\left( n \right)} = \sigma '\left( {{{\mathbf{s}}^{\left( n \right)}},{\mathbf{B}}} \right),
    \end{gathered}
\end{equation}
where ${\mathbf{q}}$ and ${{{\mathbf{s}}^{\left( n \right)}}}$ are defined in Eq.~(\ref{eq:4}). $\sigma' \left( , \right)$ is a another similarity function that measures the similarity between a vector and each column of a matrix, while it can be similar to or different from $\sigma \left( , \right)$. $\mathbf{B}$ is a feature matrix of base classes $\mathbb{C}_\textit{base}$, which can be formally represented as:
\begin{gather}
    {\mathbf{B}} = \left[ {{{\mathbf{b}}^{\left( 1 \right)}}, \ldots ,{{\mathbf{b}}^{\left( {\left| {{\mathbb{C}_\textit{base}}} \right|} \right)}}} \right].
\end{gather}
And ${{\mathbf{b}}^{\left( i \right)}}$ is the feature vector of base class $i$, which is defined as:
\begin{gather}
    {{\mathbf{b}}^{\left( i \right)}} = \frac{1}{{{M^{\left( i \right)}}}}\sum\limits_{j = 1}^{{M^{\left( i \right)}}} {GAP\left( {{{\mathbf{F}}^{\left( {i,j} \right)}}} \right)},
\end{gather}
where $M^{\left( i \right)}$ is the number of samples of base class $i$. $GAP\left(  \cdot  \right)$ is global average pooling defined in Eq.~(\ref{eq:5}). ${{{\mathbf{F}}^{\left( {i,j} \right)}}}$ is the feature map of the $j$-th sample of base class $i$ in $\mathbb{D}{_\textit{base}}$.

As can be seen from the above, Cooperative Bi-path Metric is a nonparametric (model-free) method, if we do not consider the selection of similarity functions $\sigma$, $\sigma'$ and weight hyperparameter $\alpha$. It does not introduce additional network parameters or change the training process, only additionally takes the similarity distributions of the support samples and the query samples on base classes into consideration. We can just simply append Cooperative Bi-path Metric to any trained models. However, in this way, the classification of query samples depends not only on a small number of support samples but also on the information provided by base classes, thus increasing the robustness of the model when support samples are insufficient.

\begin{algorithm}
    \caption{Cooperative Bi-path Metric with LLE}
    \label{alg:LLE}
    \LinesNumbered 
    \KwIn{${\mathbf{B}} = \left[ {{{\mathbf{b}}^{\left( 1 \right)}}, \ldots ,{{\mathbf{b}}^{\left( {\left| {{\mathbb{C}_\textit{base}}} \right|} \right)}}} \right] \in {\mathbb{R}^{c \times \left| {{\mathbb{C}_\textit{base}}} \right|}}$: feature matrix of base classes, $\mathbf{q} \in {\mathbb{R}^c}$: feature vector of query sample, $\mathbf{s}^{\left(n\right)}\in {\mathbb{R}^c}$: feature vector of novel class $n$, $k$: number of nearest neighbors to be considered for each sample, $c'$: dimensionality after reduction, ${\alpha}$: weight hyperparameter introduced in Section~\ref{sec:Cooperative Bi-path Metric}}
    \KwOut{${{\tilde \psi }^{\left( n \right)}}$: final classification score of the query sample for novel class $n$}
    \For{$i$ in $1, \ldots ,{\left| {{\mathbb{C}_\textit{base}}} \right|}$}
    {
    Find the $k$ nearest neighbors of ${\mathbf{b}}^{\left( i \right)}$ in Euclidean distance: ${{\mathbf{N}}^{\left( i \right)}} = KNN\left( {{{\mathbf{b}}^{\left( i \right)}},k} \right) = \left[ {{{\mathbf{b}}^{\left( {i,1} \right)}}, \ldots ,{{\mathbf{b}}^{\left( {i,k} \right)}}} \right]\in {\mathbb{R}^{c \times k}}$\;
    \label{step:2}
    Calculate the local covariance matrix: ${\mathbf{C}}^{\left( i \right)} = {\left( {{{\mathbf{B}}^{\left( i \right)}} - {{\mathbf{N}}^{\left( i \right)}}} \right)^ \top }\left( {{{\mathbf{B}}^{\left( i \right)}} - {{\mathbf{N}}^{\left( i \right)}}} \right)\in {\mathbb{R}^{k \times k}}$, where ${{\mathbf{B}}^{\left( i \right)}} = \left[ {{{\mathbf{b}}^{\left( i \right)}},...,{{\mathbf{b}}^{\left( i \right)}}} \right] \in {\mathbb{R}^{c \times k}}$\;
    \label{step:3}
    Calculate the weight coefficient vector: ${{\mathbf{w}}^{\left( i \right)}} = \frac{{{{\mathbf{C}}^{\left( i \right) - 1}}{{\mathbf{1}}_k}}}{{{\mathbf{1}}_k^ \top {{\mathbf{C}}^{\left( i \right) - 1}}{{\mathbf{1}}_k}}}\in {\mathbb{R}^{k}}$, where ${{\mathbf{1}}_k} \in {\mathbb{R}^k}$ is a vector full of 1 and ${{\mathbf{C}}^{\left( i \right) - 1}}$ is the inverse matrix of ${{\mathbf{C}}^{\left( i \right)}}$\;
    \label{step:4}
    }
    Calculate matrix: ${\mathbf{M}} = \left( {{\mathbf{I}}_{\left| {{\mathbb{C}_\textit{base}}} \right|} - {\mathbf{W}}} \right){\left( {{\mathbf{I}}_{\left| {{\mathbb{C}_\textit{base}}} \right|} - {\mathbf{W}}} \right)^ \top }\in {\mathbb{R}^{{\left| {{\mathbb{C}_\textit{base}}} \right|} \times {\left| {{\mathbb{C}_\textit{base}}} \right|}}}$, where $\mathbf{I}_{\left| {{\mathbb{C}_\textit{base}}} \right|} \in {\mathbb{R}^{{\left| {{\mathbb{C}_\textit{base}}} \right|} \times {\left| {{\mathbb{C}_\textit{base}}} \right|}}}$ is a identity matrix and $\mathbf{W}$'s each element
    ${{\mathbf{W}}_{j,i}} =
        \begin{cases}
            {\mathbf{w}}_k^{\left( i \right)} & \text{if ${\mathbf{b}}^{\left( j \right)}$ is the $k$-th neighbor of ${\mathbf{b}}^{\left( i \right)}$,} \\
            0                                 & \text{otherwise.}
        \end{cases}$\;
    Obtain dimensionality reduced feature matrix: ${\mathbf{\tilde B}} = \left[ {{{{\mathbf{\tilde b}}}^{\left( 1 \right)}}, \ldots ,{{{\mathbf{\tilde b}}}^{\left( {\left| {{\mathbb{C}_\textit{base}}} \right|} \right)}}} \right]\in {\mathbb{R}^{c' \times {\left| {{\mathbb{C}_\textit{base}}} \right|}}}$, where $i$-th row vector in ${\mathbf{\tilde B}}$ is the eigenvector corresponding to the $\left(i+ 1 \right)$-th smallest eigenvalue of the matrix $\mathbf{M}$\;
    \For{$\mathbf{q}$}
    {
    Find the $k$ nearest neighbors of ${\mathbf{q}}$ in ${\mathbf{B}}$ as step~\ref{step:2}: ${{\mathbf{N}}_\textit{query}} =\left[ {{{\mathbf{b}}^{\left( {1} \right)}_\textit{query}}, \ldots ,{{\mathbf{b}}^{\left( {k} \right)}_\textit{query}}} \right]\in {\mathbb{R}^{c\times k}}$ \;
    \label{step:9}
    Find $k$ corresponding dimensionality reduced vectors of ${{\mathbf{N}}_\textit{query}}$ in ${\mathbf{\tilde{B}}}$ : ${{\mathbf{\tilde{N}}}_\textit{query}} =\left[ {{{\mathbf{\tilde{b}}}^{\left( {1} \right)}_\textit{query}}, \ldots ,{{\mathbf{\tilde{b}}}^{\left( {k} \right)}_\textit{query}}} \right]\in {\mathbb{R}^{c'\times k}}$\;
    Calculate the weight coefficient vector ${{\mathbf{w}}_\textit{query}}\in {\mathbb{R}^{k}}$ as step~\ref{step:3} and~\ref{step:4} \;
    Calculate the dimensionality reduced feature vector of query sample: $\mathbf{\tilde{q}}={{\mathbf{\tilde{N}}}_\textit{query}}{{\mathbf{w}}_\textit{query}}\in {\mathbb{R}^{c'}}$\;
    \label{step:12}
    }
    \For{$\mathbf{s}^{\left(n\right)}$}
    {
        Calculate the dimensionality reduced feature vector $\mathbf{\tilde{s}}^{\left(n\right)}\in {\mathbb{R}^{c'}}$ of novel class $n$ as step~\ref{step:9}-\ref{step:12}\;
    }
    Calculate nonlinear ${{\bm{\tilde{\uprho }}}_\textit{query}}$, ${\bm{\tilde{\uprho} }}_\textit{support}^{\left( n \right)}$ and ${{\tilde \psi }^{\left( n \right)}}$ according to Eq.~(\ref{eq:6}) and (\ref{eq:7}) with dimensionality reduced ${\mathbf{\tilde B}}$, $\mathbf{\tilde{q}}$, $\mathbf{\tilde{s}}^{\left(n\right)}$ and weight hyperparameter ${\alpha}$\;
\end{algorithm}

\subsection{Revisiting Few-shot Learning with LLE}
According to Eq.~(\ref{eq:6}) and (\ref{eq:7}) in Section~\ref{sec:Cooperative Bi-path Metric}, each base class makes equal contribution to ${\varphi ^{\left( n \right)}}$, while ${{\bm{\uprho }}_\textit{query}}$ and ${\bm{\uprho }}_\textit{support}^{\left( n \right)}$ are linear about all base classes without focusing on some specific classes. Thus there arises a concern: for each query sample, different base vectors should contribute differently based on the correlations in the latent space. For example, the \textit{walker foxhound} from base classes should be prominent when querying the \textit{golden retriever} sample.


\noindent \textbf{Cooperative Bi-Path metric with LLE.} We repalce ${{\bm{\uprho }}_\textit{query}}$ and ${\bm{\uprho }}_\textit{support}^{\left( n \right)}$ with nonlinear ${{\bm{\tilde{\uprho }}}_\textit{query}}$ and ${\bm{\tilde{\uprho} }}_\textit{support}^{\left( n \right)}$ through using local linear embedding (LLE) \cite{roweis2000nonlinear}. Compared with the conventional dimensionality reduction methods such as PCA and LDA which focus on sample variance, LLE focuses on maintaining the local linear characteristics of samples when reducing sample dimensionality. LLE assumes that each sample can be represented by linearly combining its $k$ nearest neighbors, and the weight coefficient of the linear relationship before and after dimensionality reduction remains unchanged. It can be seen that LLE has some selectivity in the dimensionality reduction process, which meets our expectation that the samples should focus on some specific base classes. The process of Cooperative Bi-path Metric with LLE is shown in Alg.~\ref{alg:LLE}.

In Alg.~\ref{alg:LLE}, once ${{\bm{\tilde{\uprho }}}_\textit{query}}$, ${\bm{\tilde{\uprho} }}_\textit{support}^{\left( n \right)}$ and ${{\tilde \psi }^{\left( n \right)}}$ are obtained, query samples are assigned to novel class $\hat{y}$ with maximum classification score. Besides, Cooperative Bi-path Metric with LLE increases nonlinearity between base classes and novel samples through reducing dimensionality with LLE, thus different base classes can make different influence on classifying different query samples by the process of finding their $k$ nearest neighbors.

\section{experiments}\label{sec:4}
\subsection{Experiment Setting}

\noindent \textbf{Datasets.} We conduct experiments on two widely-used benchmarks, \ie, miniImageNet \cite{DBLP:conf/nips/VinyalsBLKW16} and tieredImageNet \cite{DBLP:conf/iclr/RenTRSSTLZ18}. MiniImageNet is a subset of ImageNet \cite{DBLP:journals/ijcv/RussakovskyDSKS15}, with 100 classes and 600 images in each class. Among these classes, 64 are for training, 16 for validation, and 20 for testing. TieredImageNet is another much larger subset of ImageNet, with 34 categories (608 classes) and 779,165 images in total. There are 20 categories (351 classes) for training, 6 categories (97 classes) for validation, and 8 categories (160 classes) for testing. Different from miniImageNet, tieredImageNet has no semantic overlap between the training set, test set, and validation set, so it is a potentially harder few-shot learning benchmark. All images are resized to $84 \times 84$ pixels.

\noindent \textbf{Training.} We follow the same experiment setting details as previous works,~\eg~\cite{DBLP:conf/nips/HouCMSC19}.
For convenience, we still organize the training data in the \emph{episodic} format for all methods, even if some methods use only query samples in the way of supervised learning. For the setting of $N$-way $K$-shot, each task samples $N$ classes from the novel classes $\mathbb{C}{_\textit{base}}$, and further samples $K$ support samples and $6$ query samples for each class. For all models, we train them for $90$ epochs on miniImageNet and $80$ epochs on tieredImageNet, and each epoch contains 1,200 tasks.

\noindent \textbf{Validation and evaluation.} After each epoch, we save the best models with maximum classification accuracy and tune the hyperparameters according to the validation set, and finally report the models' accuracy on the testing set. Similar to the training phase, the data are also organized into tasks during validation and testing phases, except that 15 query samples are sampled for each class. There are 2,000 tasks during the validation and testing phase, and we reported the models' accuracy and corresponding 95\% confidence interval on these tasks.

\noindent \textbf{Implementation details.} We implement our model using Pytorch \cite{DBLP:conf/nips/PaszkeGMLBCKLGA19}, with the part of LLE using scikit-learn \cite{scikit-learn}. All experiments are performed on a single consumer-level NVIDIA 2080Ti GPU. As mentioned in Section~\ref{sec:3.2}, we use ResNet-12 as the backbone and adopt SGD with Nesterov momentum of 0.9 and weight-decay of $5\times{10^{-4}}$ to optimize the models. The initial learning rate is set as 0.1, decreases to 0.006 at 60 epochs, and decreases to 20\% per 10 epochs thereafter. Each mini-batch contains 4 tasks during training. Except otherwise stated, the temperature hyperparameter $t$ is fixed as 0.6 for miniImageNet and 0.7 for tieredImageNet. For Cooperative Bi-path Metric, the hyperparameter values, $\alpha$, $k$, and $c'$, all take the values which produce the highest accuracy on the validation set under different settings.  The code and model are at \url{http://cvteam.net/projects/2020/CBM}.

\subsection{Comparison with the State of the Art}

To evaluate the effectiveness of the proposed approach, we first conduct experiments on the miniImageNet~\cite{DBLP:conf/nips/VinyalsBLKW16} benchmark.
As shown in Tab.~\ref{tab:1}, it can be found that although the baseline is a simple network with a fully-connected layer trained in the way of traditional supervised learning, it has achieved comparable results to the state of the art. This suggests that episodic training procedure is unnecessary and makes us rethink whether the previous complex episodic methods make sense. And we recommend this baseline for future study.
\begin{table}[]
    \caption{Comparison to prior works on 5-way classification on miniImageNet benchmark. ConvNet is a 4-layer convolutional network, WideResNet is a wider version of ResNet, baseline++ denotes our baseline with tricks and CBM (CBM+LLE) indicates Cooperative Bi-path Metric (with LLE). All the numbers of prior works are imported from corresponding original papers. The best results are bolded.}
    \label{tab:1}
    \begin{tabular}{@{}lccc@{}}
        \toprule
        Model                                                  & Backbone   & 1-shot                & 5-shot                \\ \midrule
        MAML \cite{DBLP:conf/icml/FinnAL17}                    & ConvNet    & $48.70 \pm 0.84$      & $55.31 \pm 0.73$      \\
        MN \cite{DBLP:conf/nips/VinyalsBLKW16}                 & ConvNet    & $46.6$                & $60.0$                \\
        ML-LSTM \cite{DBLP:conf/iclr/RaviL17}                  & ConvNet    & $43.44 \pm 0.77$      & $60.60 \pm 0.71$      \\
        mAP-SSVM \cite{DBLP:conf/nips/TriantafillouZU17}       & ConvNet    & $50.32 \pm 0.80$      & $63.94 \pm 0.72$      \\
        Meta-SGD \cite{DBLP:journals/corr/LiZCL17}             & ConvNet    & $50.47 \pm 1.87$      & $64.03 \pm 0.94$      \\
        Ren \emph{et al.} \cite{DBLP:conf/iclr/RenTRSSTLZ18}   & ConvNet    & $50.41 \pm 0.31$      & $64.39 \pm 0.24$      \\
        RN \cite{DBLP:conf/cvpr/SungYZXTH18}                   & ConvNet    & $50.44 \pm 0.82$      & $65.32 \pm 0.70$      \\
        Reptile \cite{DBLP:journals/corr/abs-1803-02999}       & ConvNet    & $49.97 \pm 0.32$      & $65.99 \pm 0.58$      \\
        PN \cite{DBLP:conf/nips/SnellSZ17}                     & ConvNet    & $49.92 \pm 0.78$      & $68.20 \pm 0.66$      \\
        DN4 \cite{DBLP:conf/cvpr/LiWXHGL19}                    & ConvNet    & $51.24 \pm 0.74$      & $71.02 \pm 0.64$      \\
        Gidaris \emph{et al.} \cite{DBLP:conf/cvpr/GidarisK18} & ConvNet    & $56.20 \pm 0.86$      & $73.00 \pm 0.64$      \\
        Qiao \emph{et al.} \cite{DBLP:conf/cvpr/QiaoLSY18}     & WideResNet & $59.60 \pm 0.41$      & $73.74 \pm 0.19$      \\
        SNAIL \cite{DBLP:conf/iclr/MishraR0A18}                & ResNet-12  & $55.71 \pm 0.99$      & $68.88 \pm 0.92$      \\
        TEAM \cite{DBLP:conf/iccv/Qiao000HW19}                 & ResNet-12  & $60.07$               & $75.90$               \\
        TADAM \cite{DBLP:conf/nips/OreshkinLL18}               & ResNet-12  & $58.50 \pm 0.30$      & $76.70 \pm 0.30$      \\
        CAN \cite{DBLP:conf/nips/HouCMSC19}                    & ResNet-12  & $63.85 \pm 0.48$      & $79.44 \pm 0.34$      \\
        DC \cite{DBLP:conf/cvpr/LifchitzAPB19}                 & ResNet-12  & $62.53 \pm 0.19$      & $79.77 \pm 0.19$      \\ \midrule
        baseline++ (ours)                                      & ResNet-12  & $64.07 \pm 0.45$      & $80.47 \pm 0.33$      \\
        CBM (ours)                                             & ResNet-12  & \bm{$64.77 \pm 0.46$} & $80.50 \pm 0.33$      \\
        CBM+LLE (ours)                                         & ResNet-12  & $64.21 \pm 0.45$      & \bm{$80.68 \pm 0.32$} \\ \bottomrule
    \end{tabular}
\end{table}

Also, Cooperative Bi-path Metric outperforms the baseline by a relatively large margin in the 5-way 1-shot setting, with an improvement of about 0.9\% compared to the state of the art. This indicates that the base classes help classify query samples of novel classes, especially in cases where the labeled support samples are severely insufficient.

Cooperative Bi-path Metric with LLE can slightly improve the accuracy compared with the vanilla version in the 5-way 5-shot setting, and it is about 0.9\% higher than the state of the art, which illustrates the importance of how to adaptively utilize different base classes for classifying different samples of novel classes. However, the information of base classes is underused in existing methods, we regard this as a promising but underappreciated direction in few-shot learning.

It is worth noting that Cooperative Bi-path Metric classifies query samples based on the same trained backbone of the baseline, and the accuracy improvement over the baseline is stable and not subject to the randomness of the training procedure. Compared to the baseline, Cooperative Bi-path Metric does not introduce additional network parameters or model updating processes, just changes the classification criterion during the testing phase. It is a computationally lightweight method, can be easily integrated into other trained models, such as the Prototypical Networks \cite{DBLP:conf/nips/SnellSZ17} or Matching Networks \cite{DBLP:conf/nips/VinyalsBLKW16}.

To further illustrate the effectiveness of traditional supervised learning, we also compare our baseline on tieredImageNet~\cite{DBLP:conf/iclr/RenTRSSTLZ18} with existing methods. As we can see from Tab.~\ref{tab:2}, our baseline also gets the best results in both settings.
Notably, the baseline and the proposed Cooperative Bi-path Metric (as well as the version with LLE) set up a new state of the art in the field of few-shot learning.


\subsection{Rethinking Few-shot Training Mode}


\begin{table}[]
    \caption{Comparison to prior works on 5-way classification on tieredImageNet benchmark. The best results are bolded.}
    \label{tab:2}
    \begin{tabular}{@{}lccc@{}}
        \toprule
        Model                                                & Backbone  & 1-shot                & 5-shot                \\ \midrule
        MAML \cite{DBLP:conf/icml/FinnAL17}                  & ConvNet   & $51.67 \pm 1.81$      & $70.30 \pm 1.75$      \\
        Ren \emph{et al.} \cite{DBLP:conf/iclr/RenTRSSTLZ18} & ConvNet   & $52.39 \pm 0.44$      & $69.88 \pm 0.20$      \\
        RN \cite{DBLP:conf/cvpr/SungYZXTH18}                 & ConvNet   & $54.48 \pm 0.93$      & $71.32 \pm 0.78$      \\
        PN \cite{DBLP:conf/nips/SnellSZ17}                   & ConvNet   & $53.31 \pm 0.89$      & $72.69 \pm 0.74$      \\
        CAN \cite{DBLP:conf/nips/HouCMSC19}                  & ResNet-12 & $69.89 \pm 0.51$      & $84.23 \pm 0.37$      \\ \midrule
        baseline++ (ours)                                    & ResNet-12 & \bm{$71.27 \pm 0.50$} & \bm{$85.81 \pm 0.34$} \\ \bottomrule
    \end{tabular}
\end{table}

\begin{table}[]
    \caption{Results of different training modes on 5-way classification on miniImageNet. GL: global loss; FL: few-shot loss; GL+FL: the average of global loss and few-shot loss. The best results are bolded.}
    \label{tab:3}
    \begin{tabular}{@{}lccc@{}}
        \toprule
        Model                                                     & Loss type & 1-shot                & 5-shot                \\ \midrule
        \multirow{3}{*}{TADAM \cite{DBLP:conf/nips/OreshkinLL18}} & GL        & $52.60 \pm 0.92$      & \bm{$75.10 \pm 0.67$} \\
                                                                  & FL        & $52.74 \pm 0.91$      & $73.42 \pm 0.71$      \\
                                                                  & GL+FL     & \bm{$55.97 \pm 0.93$} & $73.48 \pm 0.68$      \\ \midrule
        \multirow{3}{*}{baseline++}                               & GL        & \bm{$64.07 \pm 0.45$} & \bm{$80.47 \pm 0.33$} \\
                                                                  & FL        & $58.05 \pm 0.49$      & $74.52 \pm 0.37$      \\
                                                                  & GL+FL     & $63.15 \pm 0.47$      & $77.65 \pm 0.35$      \\ \bottomrule
    \end{tabular}
\end{table}


\noindent \textbf{Is episodic training necessary for few-shot learning?}
To make the training mode and testing mode consistent as well as to get better performance, the previous few-shot learning methods adopted the \emph{episodic} training process. More specifically, similar to the test phase, they also sampled many $N$-way $K$-shot tasks during the training phase to train a model with the cross-entropy loss of query samples over $N$ classes in each task. We call this kind of loss \emph{few-shot loss}. Different from the episodic training mode with few-shot loss, the other training mode does not only focus on the classes within a task but adopts the traditional supervised learning, using the whole training set $\mathbb{D}{_\textit{base}}$ to train a feature extractor and a ${\left| {{\mathbb{C}_\textit{base}}} \right|}$-way fully-connected layer. In this mode, the entire network is trained with the cross-entropy loss of samples over all classes in the training set. We call this kind of loss \emph{global loss}. To study the impact of different training modes on the model's accuracy, we train TADAM~\cite{DBLP:conf/nips/OreshkinLL18} and our proposed baseline in different modes. For convenience, in the experiments involving global loss, we also organize training data in the form of tasks, but only calculate query samples' global loss. For both modes, during the testing phase, the feature extractor is used to extract the features of all the samples within a task, and query samples are classified into the class with the maximum inductive similarity.

It can be seen from Tab.~\ref{tab:3} that for TADAM~\cite{DBLP:conf/nips/OreshkinLL18}, global loss alone is similar to few-shot loss alone in 1-shot setting, but it is much better than few-shot loss in 5-shot. And it is interesting to find that using both global loss and few-shot loss works worse than global loss alone in 5-shot, \emph{i.e.}, the introduction of few-shot loss reduces the performance of global loss. For baseline++, we can also find that global loss is better than few-shot loss in both settings, and the gap between themselves is larger than that for TADAM~\cite{DBLP:conf/nips/OreshkinLL18}. Through this experiment, we realize that the episodic training mode is not necessary and its capacity is limited. We also believe that using as much global information as possible to train an efficient feature extractor is important for few-shot learning.

During both training and testing phases, TADAM~\cite{DBLP:conf/nips/OreshkinLL18} uses Euclidean distance to measure the distance between samples and classes, while baseline++ uses cosine similarity. We follow the same training strategy as TADAM~\cite{DBLP:conf/nips/OreshkinLL18}, but the produced results are slightly different from the reported ones.

\subsection{Bag of Tricks for Strong Baseline}
\begin{table}[]
    \caption{Influence of the three tricks on the baseline on 5-way classification on miniImageNet benchmark. DA: data augmentation; $t$: temperature; DC: dense classification.}
    \label{tab:4}
    \begin{tabular}{@{}ccccc@{}}
        \toprule
        DA         & $t$        & DC         & 1-shot           & 5-shot           \\ \midrule
                   &            &            & $51.68 \pm 0.44$ & $69.53 \pm 0.36$ \\
        \checkmark &            &            & $59.72 \pm 0.44$ & $77.67 \pm 0.34$ \\
                   & \checkmark &            & $52.09 \pm 0.44$ & $69.80 \pm 0.37$ \\
                   &            & \checkmark & $54.00 \pm 0.45$ & $70.74 \pm 0.36$ \\
        \checkmark & \checkmark & \checkmark & $64.07 \pm 0.45$ & $80.47 \pm 0.33$ \\ \bottomrule
    \end{tabular}
\end{table}
\begin{figure}[]
    \includegraphics[width=\linewidth]{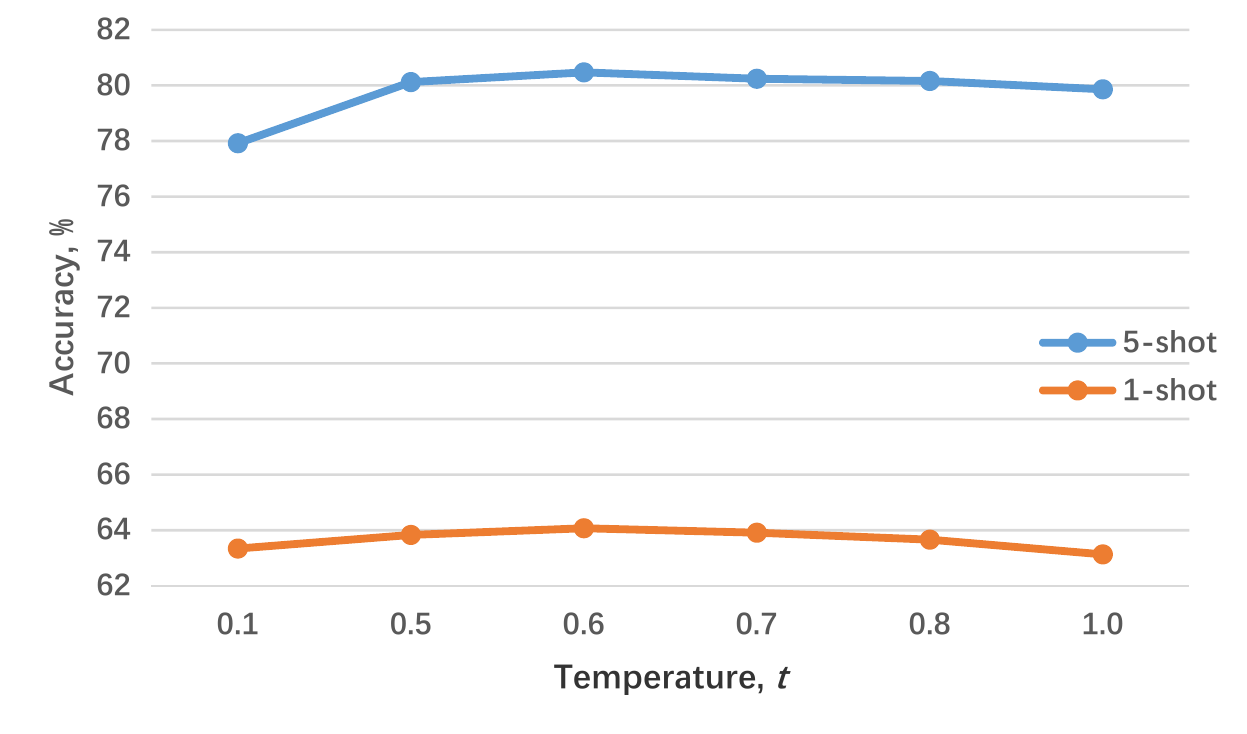}
    \caption{The accuracy curves of baseline for different values of $t$ on 5-way classification on miniImageNet benchmark. All experiments contain both data augmentation and dense classification.}
    \label{fig:3}
\end{figure}
To explore the influence of the three tricks (data enhancement, temperature, and dense classification) on the baseline, we conduct ablation experiments on miniImageNet. The experimental results are shown in Tab.~\ref{tab:4}, from which we can find that these tricks can greatly improve the accuracy of the baseline, especially data enhancement.

To further explore the impact of different values of temperature $t$ on the baseline's accuracy, we conduct experiments with different values of $t$, and the corresponding results are shown in Fig.~\ref{fig:3}. It can be seen from Fig.~\ref{fig:3} that the accuracy curves are upward convex, and the highest accuracies are obtained when $t$ is 0.6 in both settings.
All the models proposed in this paper adopt 0.6 as the value of $t$ on miniImageNet and 0.7 on tieredImageNet.
\subsection{Variants of Cooperative Bi-path Metric}
\begin{table}[]
    \caption{Accuracy of the variants of vanilla Cooperative Bi-path Metric on 5-way classification on miniImageNet benchmark. CS: cosine similarity; ED: Euclidean distance; KL: KL divergence. The best results are bolded.}
    \label{tab:5}
    \begin{tabular}{@{}ccccccc@{}}
        \toprule
        \multirow{2}{*}{$\sigma’$} & \multirow{2}{*}{softmax} & \multirow{2}{*}{$\sigma$} & \multicolumn{2}{c}{1-shot} & \multicolumn{2}{c}{5-shot}                                    \\ \cmidrule(l){4-7}
                                   &                          &                           & $\alpha$                   & Acc.                       & $\alpha$ & Acc.                  \\ \midrule
        \multirow{5}{*}{CS}        & \multirow{2}{*}{No}      & CS                        & 0.15                       & $64.60 \pm 0.46$           & 0.75     & $80.48 \pm 0.33$      \\
                                   &                          & ED                        & 0.80                       & $64.52 \pm 0.46$           & 1.00     & $80.47 \pm 0.33$      \\ \cmidrule(l){2-7}
                                   & \multirow{3}{*}{Yes}     & CS                        & 0.05                       & \bm{$64.77 \pm 0.46$}      & 0.35     & \bm{$80.50 \pm 0.33$} \\
                                   &                          & ED                        & 0.05                       & $64.75 \pm 0.46$           & 0.65     & $80.49 \pm 0.33$      \\
                                   &                          & KL                        & 0.05                       & $64.75 \pm 0.46$           & 0.50     & $80.49 \pm 0.33$      \\ \midrule
        \multirow{5}{*}{ED}        & \multirow{2}{*}{No}      & CS                        & 0.20                       & $64.62 \pm 0.46$           & 0.85     & $80.48 \pm 0.33$      \\
                                   &                          & ED                        & 0.95                       & $64.20 \pm 0.45$           & 1.00     & $80.47 \pm 0.33$      \\ \cmidrule(l){2-7}
                                   & \multirow{3}{*}{Yes}     & CS                        & 0.10                       & $64.43 \pm 0.45$           & 0.70     & $80.48 \pm 0.33$      \\
                                   &                          & ED                        & 0.10                       & $64.39 \pm 0.45$           & 0.80     & $80.48 \pm 0.33$      \\
                                   &                          & KL                        & 0.10                       & $64.45 \pm 0.45$           & 0.70     & $80.49 \pm 0.33$      \\ \bottomrule
    \end{tabular}
\end{table}
\noindent \textbf{Variants of vanilla Cooperative Bi-path Metric.} As we can see from the Tab.~\ref{tab:1}, Cooperative Bi-path Metric, as well as the version with LLE, can further improve classification accuracy by utilizing base classes during testing phase. However, we found that the specific implementation of different details of Cooperative Bi-path Metric has a considerable impact on the performance, and different variants of Cooperative Bi-path Metric have different accuracy. For the vanilla Cooperative Bi-path Metric, it needs to consider the specific implementation of these five details: (i) should cosine similarity or Euclidean distance be used as the similarity function $\sigma’$ to calculate ${{\bm{\uprho }}_\textit{query}}$ and ${\bm{\uprho }}_\textit{support}^{\left( n \right)}$ for query samples and support samples? (ii) after obtaining ${{\bm{\uprho }}_\textit{query}}$ and ${\bm{\uprho }}_\textit{support}^{\left( n \right)}$, whether softmax is applied on ${{\bm{\uprho }}_\textit{query}}$ and ${\bm{\uprho }}_\textit{support}^{\left( n \right)}$ to obtain normalized similarity distribution (all components add up to 1)? (iii) should cosine similarity or Euclidean distance or KL divergence be used as the similarity function $\sigma$ to calculate the transductive similarity ${\varphi ^{\left( n \right)}}$ between ${{\bm{\uprho }}_\textit{query}}$ and ${\bm{\uprho }}_\textit{support}^{\left( n \right)}$? (iv) how to balance the inductive similarity ${\phi ^{\left( n \right)}}$ and the transductive similarity ${\varphi ^{\left( n \right)}}$ by weight hyperparameter $\alpha$? For (i) - (iii), we enumerate and experiment with all possible combinations. For each combination, the highest accuracy is found through changing $\alpha$ in range $[0,1]$ with a interval of 0.05. All results are shown in Tab.~\ref{tab:5}. It can be found that under the setting of 1-shot, Cooperative Bi-path Metric can greatly improve the accuracy (compared with 64.07), and the value of $\alpha$ is quite small, \emph{i.e.}, the transductive similarity ${\varphi ^{\left( n \right)}}$ plays a major role. However, the result under 5-shot is not ideal. The improvement of accuracy is limited (compared with 80.47), and the value of $\alpha$ is large, so inductive similarity ${\phi ^{\left( n \right)}}$ is still in the dominant place.
\begin{table}[]
    \caption{Accuracy of the variants of Cooperative Bi-path Metric with LLE on 5-way classification on miniImageNet benchmark. The dimensionality after reduction $c'$ is fixed as 63. The best results are bolded.}
    \label{tab:6}

    \scalebox{0.8}
    {
        \begin{tabular}{@{}cccccccccc@{}}
            \toprule
            \multirow{2}{*}{$L_2$} & \multirow{2}{*}{$\sigma’$} & \multirow{2}{*}{softmax} & \multirow{2}{*}{$\sigma$} & \multicolumn{3}{c}{1-shot} & \multicolumn{3}{c}{5-shot}                                                                  \\ \cmidrule(l){5-10}
                                   &                            &                          &                           & $k$                        & $\alpha$                   & Acc.                  & $k$ & $\alpha$ & Acc.                  \\ \midrule
            \multirow{10}{*}{No}   & \multirow{5}{*}{CS}        & \multirow{2}{*}{No}      & CS                        & 8                          & 0.95                       & $64.16 \pm 0.45$      & 24  & 0.95     & \bm{$80.68 \pm 0.32$} \\
                                   &                            &                          & ED                        & 8                          & 0.95                       & $64.17 \pm 0.45$      & 24  & 0.95     & $80.67 \pm 0.32$      \\ \cmidrule(l){3-10}
                                   &                            & \multirow{3}{*}{Yes}     & CS                        & 10                         & 0.35                       & \bm{$64.21 \pm 0.45$} & 26  & 0.35     & $80.58 \pm 0.32$      \\
                                   &                            &                          & ED                        & 11                         & 0.30                       & $64.19 \pm 0.45$      & 24  & 0.30     & $80.59 \pm 0.32$      \\
                                   &                            &                          & KL                        & 10                         & 0.35                       & $64.20 \pm 0.45$      & 25  & 0.25     & $80.63 \pm 0.32$      \\ \cmidrule(l){2-10}
                                   & \multirow{5}{*}{ED}        & \multirow{2}{*}{No}      & CS                        & 4                          & 0.25                       & $64.14 \pm 0.45$      & 23  & 0.10     & $80.55 \pm 0.32$      \\
                                   &                            &                          & ED                        & 5                          & 1.00                       & $64.07 \pm 0.45$      & 63  & 0.95     & $80.49 \pm 0.32$      \\ \cmidrule(l){3-10}
                                   &                            & \multirow{3}{*}{Yes}     & CS                        & 21                         & 0.55                       & $64.11 \pm 0.45$      & 33  & 0.60     & $80.50 \pm 0.32$      \\
                                   &                            &                          & ED                        & 22                         & 0.55                       & $64.11 \pm 0.45$      & 23  & 0.70     & $80.51 \pm 0.32$      \\
                                   &                            &                          & KL                        & 22                         & 0.50                       & $64.11 \pm 0.45$      & 23  & 0.40     & $80.51 \pm 0.32$      \\ \midrule
            \multirow{10}{*}{Yes}  & \multirow{5}{*}{CS}        & \multirow{2}{*}{No}      & CS                        & 7                          & 0.95                       & $64.20 \pm 0.45$      & 23  & 0.95     & $80.64 \pm 0.32$      \\
                                   &                            &                          & ED                        & 8                          & 0.95                       & $64.20 \pm 0.45$      & 22  & 0.95     & $80.63 \pm 0.32$      \\ \cmidrule(l){3-10}
                                   &                            & \multirow{3}{*}{Yes}     & CS                        & 5                          & 0.30                       & $64.19 \pm 0.45$      & 12  & 0.55     & $80.57 \pm 0.32$      \\
                                   &                            &                          & ED                        & 9                          & 0.40                       & $64.18 \pm 0.45$      & 25  & 0.30     & $80.58 \pm 0.32$      \\
                                   &                            &                          & KL                        & 7                          & 0.30                       & $64.20 \pm 0.45$      & 23  & 0.30     & $80.61 \pm 0.32$      \\ \cmidrule(l){2-10}
                                   & \multirow{5}{*}{ED}        & \multirow{2}{*}{No}      & CS                        & 16                         & 0.10                       & $64.16 \pm 0.45$      & 25  & 0.10     & $80.58 \pm 0.32$      \\
                                   &                            &                          & ED                        & 5                          & 1.00                       & $64.07 \pm 0.45$      & 51  & 0.95     & $80.65 \pm 0.32$      \\ \cmidrule(l){3-10}
                                   &                            & \multirow{3}{*}{Yes}     & CS                        & 5                          & 0.45                       & $64.12 \pm 0.45$      & 23  & 0.75     & $80.51 \pm 0.32$      \\
                                   &                            &                          & ED                        & 5                          & 0.40                       & $64.12 \pm 0.45$      & 20  & 0.65     & $80.51 \pm 0.32$      \\
                                   &                            &                          & KL                        & 7                          & 0.30                       & $64.12 \pm 0.45$      & 22  & 0.70     & $80.51 \pm 0.32$      \\ \bottomrule
        \end{tabular}
    }
\end{table}

\noindent \textbf{Variants of Cooperative Bi-path Metric with LLE.} For Cooperative Bi-path Metric with LLE, it requires additional determination of (v) whether or not to conduct $L_\textit{2}$ normalization on feature vectors before LLE and the values of two additional hyperparameters, namely (vi) the number of nearest neighbors $k$ and (vii) the dimensionality after reduction $c'$ in LLE. Like vanilla Cooperative Bi-path Metric, for (i) - (vi), we enumerate and experiment with all possible combinations. For (vii), through some preliminary experiments, we find that the highest accuracy of various combinations are obtained when $c'$ is 63, so we fix $c'$ as 63 for all combinations. And the highest accuracy under various combinations is shown in Tab.~\ref{tab:6}.
\begin{figure}[]
    \includegraphics[width=\linewidth]{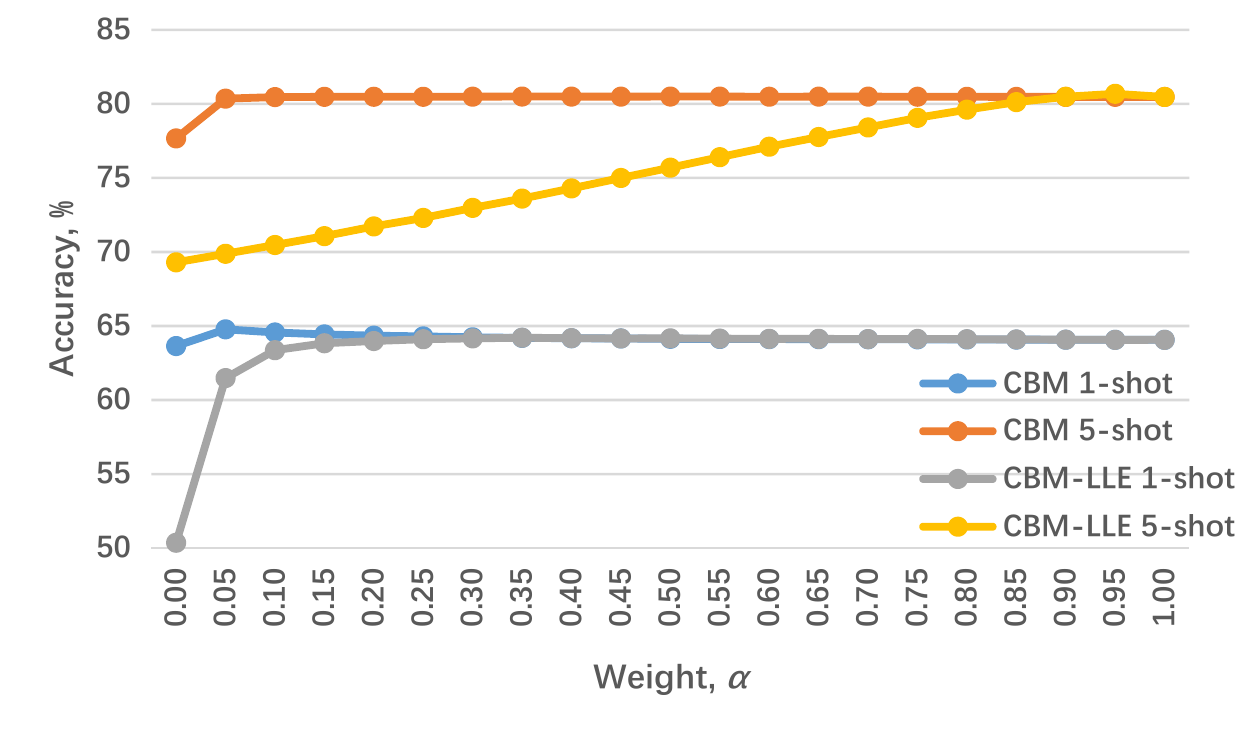}
    \caption{The accuracy curves for different values of $\alpha$ on 5-way classification on miniImageNet benchmark.}
    \label{fig:4}
\end{figure}

\noindent \textbf{Influence of weight hyperparameter.} In order to further explore the influence of weight hyperparameter $\alpha$ on classification accuracy (especially when $\alpha=0$, only transductive similarity ${\varphi ^{\left( n \right)}}$ works), we report the accuracy curves with different values of $\alpha$, as shown in Fig.~\ref{fig:4}. It can be seen from Fig.~\ref{fig:4}, when ${\phi ^{\left( n \right)}}$ is the majority, the accuracy is higher, ${\phi ^{\left( n \right)}}$ contributes more to the classification than ${\varphi ^{\left( n \right)}}$. We attribute part of the reason why ${\phi ^{\left( n \right)}}$ takes a larger part to that the magnitude of ${\phi ^{\left( n \right)}}$ and ${\varphi ^{\left( n \right)}}$ is different. The accuracy at $\alpha=0$ is less than the accuracy at $\alpha=1$, which means that classification result based on ${\varphi ^{\left( n \right)}}$ alone is less accurate than that based on ${\phi ^{\left( n \right)}}$ alone. However, the highest points of the accuracy curves are not obtained at $\alpha=1$, so ${\varphi ^{\left( n \right)}}$ can further improve the classification accuracy based on ${\phi ^{\left( n \right)}}$.

\section{Conclusions and future work}
In this work, we contribute to few-shot learning with a concise and effective baseline as well as a novel metric named Cooperative Bi-path Metric.

First, we train a simple network in the way of traditional supervised learning as the baseline, which achieves comparable results to the state of the art. This shows that episodic training mode is not necessary, and an effective feature extractor to capture discriminative features of samples is fundamental for few-shot learning.

Second, we propose Cooperative Bi-path Metric to change the criterion of classification. It uses samples' similarity distribution on base classes to assist classification decisions while existing methods did not take full use of such information of base classes. Experiments show that it can further boost the model's performance and achieve a new state of the art in the field of few-shot image classification, indicating that using base classes to classify samples during the testing phase looks like a promising direction for future research.

However, Cooperative Bi-path Metric is handcrafted and somewhat straightforward. A natural direction for improving it is training an additional convolutional neural network end to end to measure the transductive similarity. We leave this for future work.
\begin{acks}
    This work was supported in part by the National Key R\&D Program of China (No. 2017YFB1002400), the National Natural Science Foundation of China (No. 61922006, No. 61532003, No. 61825101 and No. U1611461), and the Beijing Nova Program (No. Z181100006218063).
\end{acks}
\bibliographystyle{ACM-Reference-Format}
\balance
\bibliography{sample-base}
\end{document}